\documentclass[letterpaper, 10 pt, conference]{ieeeconf}  

\IEEEoverridecommandlockouts                              
\usepackage[utf8]{inputenc}
\usepackage[T1]{fontenc}

\usepackage{amsmath} 
\usepackage{amssymb}  
\usepackage{algorithm}
\usepackage{algorithmic}
\usepackage{float}
\usepackage[section]{placeins}

\usepackage{graphicx}
\usepackage{url}
\usepackage{caption}
\title{\LARGE \bf
One-Step Model Predictive Path Integral for Manipulator Motion Planning Using Configuration Space Distance Fields
}

\author{Yulin Li$^{1}$, Tetsuro Miyazaki$^{1}$ and Kenji Kawashima$^{1}$
\thanks{*This work was supported by JSPS KAKENHI Grant Numbers 25H00717.}
\thanks{$^{1}$Department of Information Physics and Computing, 
        The University of Tokyo, 7-3-1 Hongo, Bunkyo-Ku, Tokyo, Japan
        {\tt\small kenji-kawashima@ipc.i.u-tokyo.ac.jp}}%
}

\begin{document}

\maketitle
\thispagestyle{empty}
\pagestyle{empty}

\begin{abstract}

Motion planning for robotic manipulators is a fundamental problem in robotics. 
Classical optimization-based methods typically rely on the gradients of signed distance fields (SDF) to impose collision-avoidance constraints. 
However, these methods are susceptible to local minima and may fail when the SDF gradients vanish. Recently, Configuration Space Distance Fields (CDFs) have been introduced, which directly model distances in the robot’s configuration space. 
Unlike workspace SDF, CDFs are differentiable almost everywhere and thus provide reliable gradient information. On the other hand, gradient-free approaches such as Model Predictive Path Integral (MPPI) control leverage long-horizon rollouts to achieve collision avoidance. While effective, these methods are computationally expensive due to the large number of trajectory samples, repeated collision checks, and the difficulty of designing cost functions with heterogeneous physical units. In this paper, we propose a framework that integrates the CDF representation with MPPI to enable direct navigation in the robot’s configuration space. Leveraging CDF gradients, we unify the MPPI cost in joint space and reduce the horizon to one step, substantially cutting computation while preserving collision avoidance in practice. We demonstrate that our approach achieves nearly 100\% success rates in 2D environments and consistently high success rates in challenging 7-DoF Franka manipulator simulations with complex obstacles. 
Furthermore, our method attains control frequencies exceeding 750 Hz, substantially outperforming both optimization-based and standard MPPI baselines. 
These results highlight the effectiveness and efficiency of the proposed CDF-MPPI framework for high-dimensional motion planning.

\end{abstract}


\section{Introduction}
Motion planning is one of the most active research areas in robotic control, with extensive efforts devoted to achieving efficient and collision-free robot motions. Many studies \cite{intro_1,intro_2} formulate this problem as an optimization problem, where the signed distance field (SDF) is used to represent obstacle proximity and its gradient is employed to guide the optimization toward collision-free trajectories. In practice, classical approaches such as CHOMP \cite{intro_2} precompute SDF on a voxel grid via the Euclidean Distance Transform (EDT), while TrajOpt \cite{intro_1} repeatedly queries distances using Gilbert–Johnson–Keerthi (GJK) \cite{GJK} between convex bodies. These computations are often implemented on CPUs in standard libraries, which is generally sufficient for static environments and moderate-scale problems. Recent work has explored GPU acceleration of both SDF construction and GJK queries to handle dynamic scenes and large-scale planning more efficiently, though they can still be computationally expensive due to frequent voxel updates and the iterative nature of distance queries.

More recent works, such as \cite{sdf2_li2024representing}, which adopt regression-based formulations, and \cite{sdf1_koptev2022neural}, which leverage neural networks, demonstrate that SDF can be efficiently fitted and evaluated on GPU. These approaches enable faster and more accurate SDF computation. However, as in previous studies, gradient information is still required to achieve collision-free motion. For robotic arms, this introduces two issues: (i) the SDF gradient may vanish near obstacles \cite{gradient_prob}, and (ii) the distribution of SDF values in the joint space is highly non-uniform \cite{cdf}.

\begin{figure}[t]
    \centering
    \includegraphics[width=\linewidth]{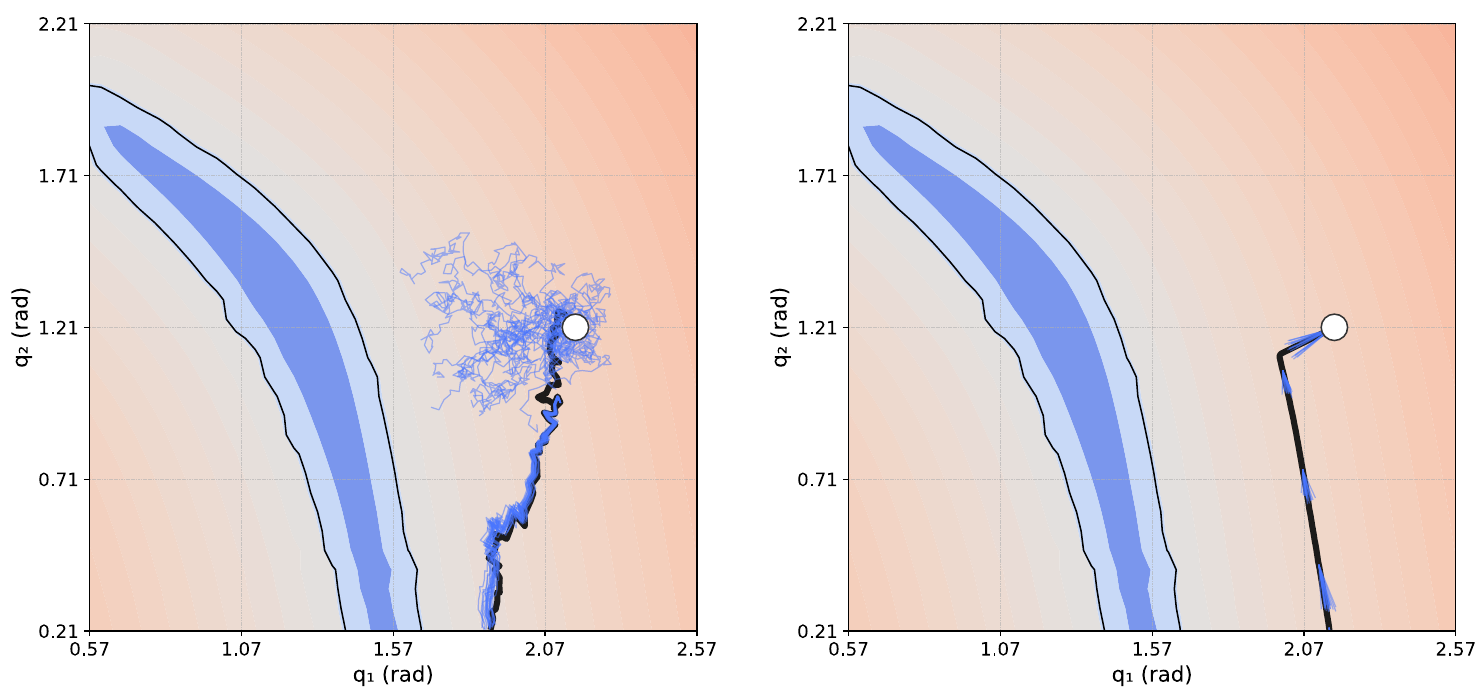}
    \caption{Comparison between the original MPPI and our proposed method in the two-link robot configuration space. 
    The irregular shapes on the left represent obstacles, blue thin lines denote sampled trajectories at each step, the black line indicates the resulting trajectory, and the white dot marks the start configuration. 
    \textbf{Left:} the original MPPI samples long-horizon trajectories at each step. 
    \textbf{Right:} our method leverages the CDF to sample one-step horizon trajectories at each step.}
    \label{story_1}
\end{figure}

To overcome these limitations, sampling-based methods have emerged as a viable alternative. Their key advantage lies in the fact that they do not require gradient information, thereby avoiding issues such as the vanishing-gradient problem in SDF. Representative approaches include Rapidly-exploring Random Trees (RRT) \cite{RRT} and Probabilistic Roadmaps (PRM) \cite{PRM}. However, these methods are often inefficient due to their reliance on extensive random sampling. 

Model Predictive Control (MPC) has also been widely adopted in robotic control~\cite{MPC_1,MPC_2}. However, incorporating obstacle avoidance typically leads to highly non-convex optimization problems, making optimization-based MPC solvers sensitive to initialization and prone to local minima, especially in high-dimensional and cluttered configuration spaces.

Model Predictive Path Integral (MPPI) control was introduced in \cite{mppi}, integrating the sampling paradigm with the predictive principles of MPC. By updating the sampling distribution iteratively, MPPI improves sampling efficiency while eliminating the need for optimization solvers. Its extension to robotic manipulators was presented in \cite{sampling_mpc}, demonstrating promising results. In this framework, collision-free trajectories are obtained by penalizing predicted collisions within the cost function, rather than relying on gradient information, thereby mitigating the vanishing-gradient issue. Nevertheless, two key challenges remain. First, the cost function design in \cite{sampling_mpc} is intricate, as it involves heterogeneous components that are not directly comparable. Second, the method is computationally expensive: at each time step, a large number of finite-horizon trajectory rollouts must be simulated, each requiring frequent collision checks with obstacles. This significantly increases computation time and imposes heavy demands on GPU resources \cite{sdf1_koptev2022neural}.

Configuration Space Distance Fields (CDFs), proposed in \cite{cdf}, provide a promising alternative. Unlike traditional SDF defined in the workspace, CDFs directly construct distance fields in the configuration space, thereby addressing both the non-uniformity and vanishing-gradient problems. CDFs thus offer an attractive framework for robot motion planning, as their gradients can be directly leveraged for navigation in configuration space \cite{cdf}. However, optimization-based methods that rely on CDFs are susceptible to local minima, which may result in navigation failures. Moreover, such optimization problems typically require repeated iterations, making the computation slow especially in high-dimensional configuration spaces.

In this work, we integrate the MPPI framework with the CDF representation and propose a simple yet efficient algorithm that addresses several limitations of existing methods. Our key idea is to simplify MPPI’s sampling by reducing its reliance on long-horizon rollouts, which are typically used to implicitly explore obstacles and accumulate future costs. Instead, we use the CDF representation as a direct provider of local geometric safety cues in configuration space, namely distance and directional information, so collision risk and goal progress can be evaluated locally. This allows us to operate MPPI with short-horizon control sampling at a much lower computational cost.
\begin{enumerate}
    \item \textbf{Avoiding local optima:} By eliminating optimization-based procedures, our method reduces the risk of convergence to local minima.
    \item \textbf{Unified and simplified cost function:} Leveraging the property that CDFs provide well-distributed, non-vanishing gradients in the configuration space, we redesign the MPPI cost function. In particular, we employ angular navigation to normalize and unify the scale of different cost terms.
    \item \textbf{Faster computation:} The redesigned cost function enables one-step trajectory sampling instead of requiring long-horizon rollouts, thereby significantly reducing computation time. Most computations can be efficiently executed on GPUs (Fig.~\ref{story_1} illustrates a comparison between our method and the original MPPI).
\end{enumerate}

\section{PRELIMINARIES}

\subsection{Model Predictive Path Integral}
The MPPI framework operates as follows. For a discrete-time robotic system at time step $t$, the control input ${\mathbf{u}}$ represents the joint velocity. The input ${\mathbf{u}}$ is sampled from a policy 
$\pi_t = \prod_{h=0}^{H-1} \pi_{t, h}$
where $H$ is the prediction horizon and $\pi_{t, h}$ is a Gaussian distribution with mean $\boldsymbol{\mu}_{t, h}$ and covariance $\boldsymbol{\Sigma}_{t, h}$.  

At each time step $t$, the MPPI algorithm proposed by \cite{sampling_mpc} draws a batch of $N$ control sequences 
${{\mathbf{u}}}_{i=1..N}^{h=1..H}$ from the current policy $\pi_t$.  
For each sampled sequence, the corresponding cost ${c_{i,h}}$ is computed from the state trajectories ${x}_{i,h}$ obtained by propagating the dynamics of the robot using the sampled controls.  

The Gaussian policy parameters $\pi_{t, h} = \mathcal{N}(\boldsymbol{\mu_{t,h}}, {\boldsymbol{\Sigma}}_{t,h})$ are then updated according to the following weighted rules:
\begin{subequations}
\begin{align}
    \boldsymbol{\mu}_{t + 1,h} \leftarrow & (1 - \alpha_{\boldsymbol{\mu}}){\boldsymbol{\mu}}_{t,h} 
    + \alpha_{\boldsymbol{\mu}} \frac{\sum_{i=1}^N w_i \mathbf{u}_{i,h}}{\sum_{i=1}^N w_i}, \label{mean_update} \\[6pt]
    {\boldsymbol{\Sigma}}_{t + 1,h} \leftarrow & (1 - \alpha_{\boldsymbol{\Sigma}}){\boldsymbol{\Sigma}}_{t,h} \nonumber \\
    &+ \alpha_{\boldsymbol{\Sigma}} 
    \frac{\sum_{i=1}^N w_i (\mathbf{u}_{i,h} - {\boldsymbol{\mu}}_{t,h})(\mathbf{u}_{i,h} - {\boldsymbol{\mu}}_{t,h})^\top}
         {\sum_{i=1}^N w_i}.
    \label{cov_update}
\end{align}
\end{subequations}
where $\alpha_{\boldsymbol{\mu}}$ and $\alpha_{\boldsymbol{\Sigma}}$ are filtering coefficients. The weight for each sample is defined as
$w_i = \exp\left( -\frac{1}{\beta} \hat{C}_i \right)$,
where $\beta$ is a temperature parameter for the cost.  
The cost value $\hat{C}(\mathbf{x}_t, {\mathbf{u}})$ is computed as
\begin{equation}
    \hat{C}(\mathbf{x}_t, {\mathbf{u}}) = \sum_{h=0}^{H-2} \gamma^h c(\mathbf{x}_{i,h}, \mathbf{u}_{i,h}) + \gamma^{H-1} \hat{c}(\mathbf{x}_{i,H-1}, \mathbf{u}_{i,H-1}), \label{cost}
\end{equation}
where $\gamma$ is the discount factor, $c(\cdot)$ is the stage cost, $\hat{c}(\cdot)$ denotes the terminal cost, and $i$ indexes each trajectory. This weighting scheme biases the sampling toward lower-cost control sequences.  

The stage cost $c_{\mathbf{x},{\mathbf{u}}}$ is composed of multiple terms, including joint-limit avoidance, self-collision and collision avoidance, contingency stopping, and manipulation cost:
\begin{equation}
    c_{\mathbf{x}, {\mathbf{u}}} = \alpha_s c_{stop} + \alpha_m c_{mani} + \alpha_c c_{col} + \alpha_j c_{joint}. \label{cost_mpc}
\end{equation}
Here, $\alpha_s, \alpha_m, \alpha_c$, and $\alpha_j$ are weighting coefficients for the respective cost components.  
Further details can be found in \cite{sampling_mpc}.

\subsection{Configuration Space Distance Fields}
We first define the SDF of the robot, denoted by $f_s$, as a function that assigns a signed distance between a point in the workspace $\mathbf{p}$ and the robot at configuration $\mathbf{q}$:
\begin{equation}
    f_s(\mathbf{p}, \mathbf{q}) = \pm \min_{\mathbf{p}' \in \mathcal{R}(\mathbf{q})} \, \| \mathbf{p} - \mathbf{p}' \|,
\end{equation}
where $\mathcal{R}(\mathbf{q})$ represents the set of points occupied by the robot at configuration $\mathbf{q}$, and $\mathbf{p}'$ is the closest point on the robot to $\mathbf{p}$. The CDF, denoted by $f_c$, measures the minimum distance from a robot configuration $\mathbf{q}$ to the set of configurations that are in contact with obstacles in the workspace~\cite{cdf}:
\begin{equation}
    f_c(\mathbf{p}, \mathbf{q}) = \min_{\mathbf{q}'} \, \| \mathbf{q} - \mathbf{q}' \|,
\end{equation}
subject to the constraint that $f_s(\mathbf{p}, \mathbf{q}') = 0$,  
where $\mathbf{q}'$ denotes a configuration in contact with the obstacle and $\mathbf{p}$ is a corresponding contact point in the workspace. In practice, $f_c(\mathbf{p}, \mathbf{q})$ can be encoded using a neural network. The CDF representation possesses several desirable properties. In particular, it satisfies the eikonal equation $\| \nabla_{\mathbf{q}} f_c(\mathbf{p}, \mathbf{q}) \| = 1$ almost everywhere in the configuration space.  
This property makes CDFs especially well-suited for motion planning, since the gradient $\nabla_{\mathbf{q}} f_c$ naturally points in the locally optimal escape direction away from the nearest obstacle, thereby providing a direct guidance signal for collision avoidance.

\section{Method}
We aim to control an $n$-DOF robotic arm to perform fast, collision-free motions.
In this section, we present a method that combines MPPI and the CDF representation to achieve this goal.

The robot dynamics are modeled as:
\begin{equation}
    \mathbf{q}_{t+1} = \mathbf{q}_t + \Delta t \cdot \mathbf{u} .
    \label{dynamic}
\end{equation}
where $\mathbf{q}_t$ is the joint configuration at time step $t$, $\mathbf{u}$ is the joint input, and $\Delta t$ is the discrete control time step.

\subsection{Model Predictive
Path Integral with an Angle-Based Cost Function}
We propose an \emph{angle-based} cost function that unifies the cost terms into a common unit. Let $\mathbf{q}_t$ be the robot configuration at time $t$, $\mathbf{q}_{t+1}$ the configuration at the next time step, $\mathbf{q}_f$ the goal configuration, and $\mathbf{p}$ is the obstacle point in the workspace.  
The proposed cost is defined as:
\begin{equation}
    c(\theta_1, \theta_2) = \alpha_1 \theta_1 + \alpha_2 \theta_2 , \label{cost_our_method}
\end{equation}
where $\alpha_1, \alpha_2 > 0$ are scalar weights. $\theta_1 \in [0, \pi]$ is the angle between the obstacle normal direction at $\mathbf{q}_t$ and the motion vector $\mathbf{q}_{\mathrm{next}} = (\mathbf{q}_{t+1} - \mathbf{q}_t)$, while $\theta_2 \in [0, \pi]$ is the angle between the goal vector $\mathbf{e}_{t} = (\mathbf{q}_f - \mathbf{q}_t)$ and the motion vector $\mathbf{q}_{\mathrm{next}}$.  
When $\theta_2 = 0$, the robot moves directly toward the goal; $\theta_2 = \pi$ indicates motion directly away from it. The angle $\theta_2$ can be computed as:
\begin{equation}
\begin{aligned} 
    \theta_2(\mathbf{q}_t, \mathbf{q}_{t + 1}) &= \arccos \frac{\mathbf{q}_{next} \cdot \mathbf{e}_{t}}
         {\| \mathbf{q}_{next} \| \, \| \mathbf{e}_{t} \| },
\end{aligned}
\label{obs_angle}
\end{equation}
For the $\theta_1$, we calculated from
\begin{equation}
\theta_1(\mathbf{q}_t, \mathbf{q}_{t + 1}) =
\begin{cases}
0, &
\begin{aligned}
\text{if } & \theta_1^\star < \tfrac{\pi}{2} \ \text{or} \\
           &  f_c(\mathbf{p}, \mathbf{q}_t) \ge d_{\mathrm{act}} \ \text{or} \\
           & f_c(\mathbf{p}, \mathbf{q}_t) \ge \lVert \mathbf{e}_{t} \rVert ,
\end{aligned} \\[4pt]
\theta_1^\star, & \text{otherwise.}
\end{cases}
\label{obs_cost_1}
\end{equation}

\begin{equation}
\theta_1^\star =
\arccos \frac{\mathbf{q}_{\mathrm{next}} \cdot 
\nabla_{\mathbf{q}} f_c(\mathbf{p}, \mathbf{q}_t)}
{\lVert \mathbf{q}_{\mathrm{next}} \rVert \,
 \lVert \nabla_{\mathbf{q}} f_c(\mathbf{p}, \mathbf{q}_t) \rVert},
\label{obs_cost_2}
\end{equation}

\begin{figure}[t]
    \centering
    \includegraphics[width=\linewidth]{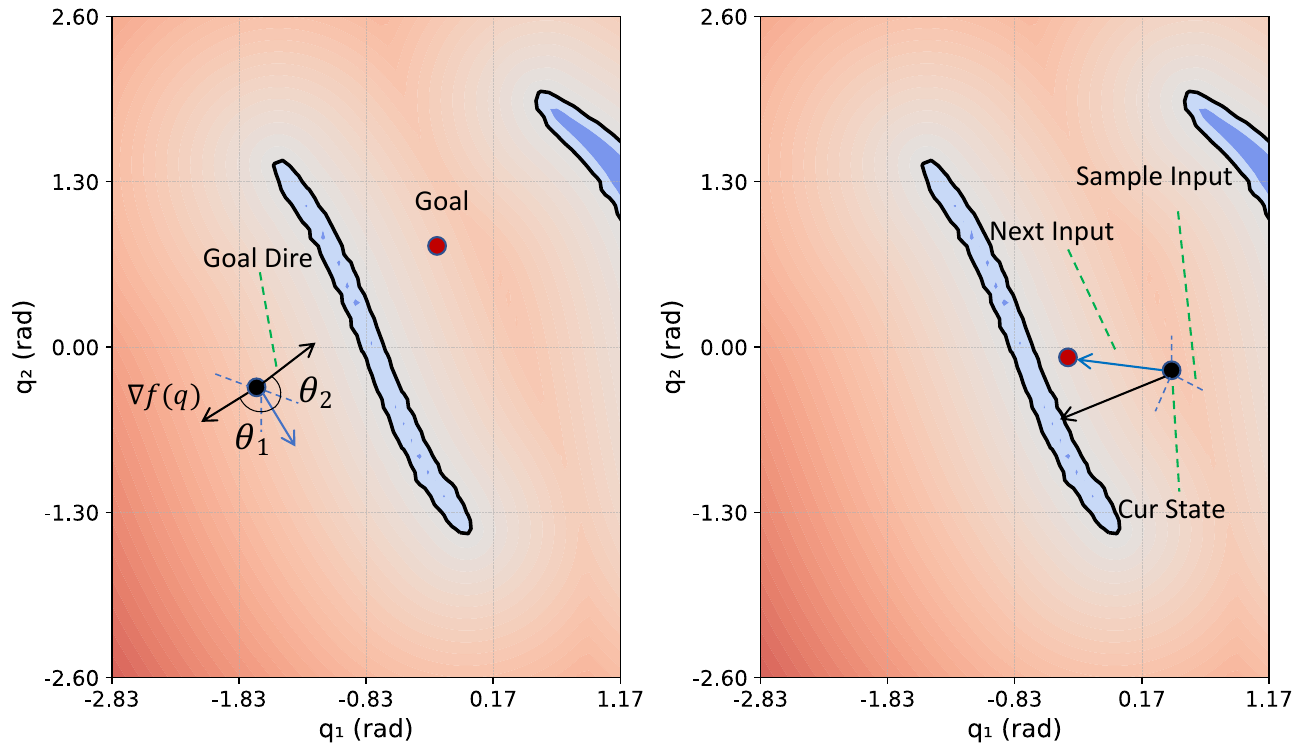}
    \caption{%
   Two-link robot in the configuration space. Irregular shapes represent obstacles, and the black dot denotes the current state of the robot. 
    \textbf{Left}: The blue arrow indicates the actual motion direction of the robot at the next step, forming two angles with the CDF gradient and the goal state. 
    \textbf{Right}: Illustration showing why the obstacle-avoidance angle cost is set to zero when the robot--obstacle distance exceeds the robot--goal distance.}
    \label{ill_1}
\end{figure}

\begin{figure}[t]
    \centering
    \includegraphics[width=\linewidth]{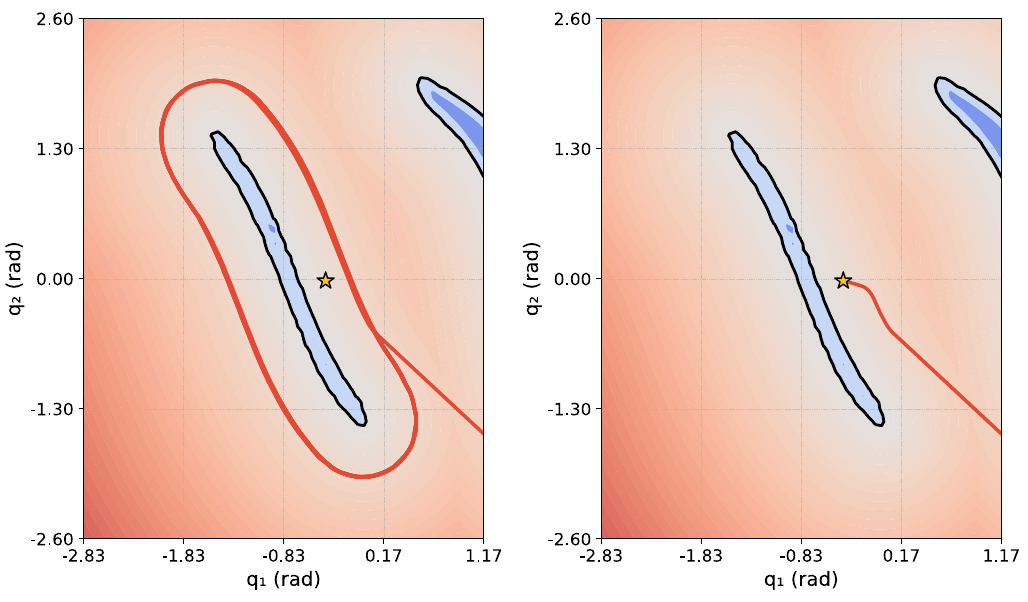}
    \caption{
    \textbf{Left}: If the robot–obstacle distance exceeds the robot–goal distance while the obstacle cost remains active, the robot may circle around the obstacle. 
    \textbf{Right}: When the obstacle cost is set to zero under the same condition, the robot converges to the goal more directly. 
    The star marker indicates the goal configuration.
    }
    \label{ill_2}
\end{figure}

The obstacle normal direction is obtained from the gradient of the collision CDF, 
$\nabla_{\mathbf{q}} f_c(\mathbf{p}, \mathbf{q}_t) $, which points towards the optimal escape direction from the nearest obstacle. 
In our implementation, $\theta_1$ is set to zero under the following conditions:  

\begin{itemize}
    \item \textbf{Angle condition:} If $\theta_1^\star < \pi / 2$, the motion naturally moves away from the obstacle, and collision will not occur (see Eq.~\ref{obs_cost_1} and Fig.~\ref{ill_1}(Left)).  
    \item \textbf{Distance threshold:} If the distance to the obstacle $f_c(\mathbf{p}, \mathbf{q}_t)$ exceeds a specified activation threshold $d_{act}$ obtained from the CDF, then obstacle avoidance is unnecessary because the robot is already far from the obstacle.  
    \item \textbf{Goal proximity:} If the distance to the obstacle is larger than the distance to the goal, i.e., $f_c(\mathbf{p}, \mathbf{q}_t) \ge \lVert \mathbf{e}_{t} \rVert$, as illustrated in Fig.~\ref{ill_1}(Right) and Fig.~\ref{ill_2}, it is more efficient to move directly towards the goal rather than proactively avoiding the obstacle.  
\end{itemize}

This design offers two advantages:  
(1) all cost terms are expressed in the same unit (angle), making the weight selection more intuitive;  
(2) obstacle avoidance and goal-seeking behaviors are directly captured in a single unified metric. In MPPI, a batch of $N$ control sequences is sampled from $\pi_{t}$, and each sequence is rolled out over a prediction horizon $H$ to compute the total cost.  
With the proposed angle-based cost, collision risk and goal progress can be evaluated locally at the current time step, without simulating the full horizon.  
Therefore, we modify the sampling policy to generate only $N$ single-step control inputs ($H = 1$) from $\pi_{t}$, compute $\mathbf{q}_{next}$, and evaluate the cost directly.  
This reduces computation while retaining effective obstacle avoidance and goal-directed motion. 

For the remaining cost terms in Eq.~\ref{cost_mpc}, the manipulability term $c_{\text{mani}}$ is omitted, as our planning is conducted directly in the configuration space and thus does not suffer from the low-manipulability issue reported in \cite{sdf2_li2024representing}. 
For the collision cost $c_{\text{col}}$, we also treat the robot itself as an obstacle in the CDF function to account for self-collisions. 
For the joint-limit cost $c_{\text{joint}}$, we employ a simple projection strategy to avoid joint-limit violations, which will be detailed in the next section. 
With these considerations, the cost function in Eq.~\ref{cost} can be rewritten as
\begin{equation}
    \hat{C}(\mathbf{x}_t, \mathbf{u}) = {c}(\mathbf{x}_{i}).
\end{equation}
where $c$ is defined in Eq.~\ref{cost_our_method}.

\subsection{One-step MPPI for Configuration Space Motion Planning}

To handle joint limits, we simply project $\mathbf{u}$ into feasible control set,
\begin{equation}
\mathcal{U}_{\text{step}}(\mathbf{q})
=\bigl\{\mathbf{u}\;\big|\; 
\mathbf{q}+\Delta t\,\mathbf{u}\in[\mathbf{q}_{\min},\mathbf{q}_{\max}]\bigr\}.
\end{equation}
Equivalently, this yields the elementwise bounds
\begin{equation}
\mathbf{u}_{\min}(\mathbf{q})=\frac{\mathbf{q}_{\min}-\mathbf{q}}{\Delta t},
\qquad
\mathbf{u}_{\max}(\mathbf{q})=\frac{\mathbf{q}_{\max}-\mathbf{q}}{\Delta t}.
\label{eq:step-u-bounds}
\end{equation}
The projection of any control $\mathbf{u}\in\mathbb{R}^n$ onto 
$\mathcal{U}_{\text{step}}(\mathbf{q})$ is the elementwise clipping
\begin{equation}
\mathrm{proj}_{\mathcal{U}_{\text{step}}(\mathbf{q})}(\mathbf{u})
=\operatorname{clip}\!\bigl(\mathbf{u};\,
\mathbf{u}_{\min}(\mathbf{q}),\,\mathbf{u}_{\max}(\mathbf{q})\bigr),
\label{eq:proj-step-box}
\end{equation}
i.e., for each coordinate $i=1,\dots,n$,
\begin{equation}
\bigl[\mathrm{proj}_{\mathcal{U}_{\text{step}}(\mathbf{q})}(\mathbf{u})\bigr]_i
=\min\!\Bigl\{\max\!\bigl(u_i,\ [\mathbf{u}_{\min}(\mathbf{q})]_i\bigr),\
[\mathbf{u}_{\max}(\mathbf{q})]_i\Bigr\}. \label{projection}
\end{equation}

We have discussed the main idea of our method in the previous sections.
The complete algorithm is presented in Algorithm~\ref{ourMethod}. Regarding the choice of hyperparameters, the relative magnitudes of $\alpha_1$ and $\alpha_2$ determine the trade-off between obstacle avoidance and goal-seeking.  
If a safer behavior is desired, $\alpha_1$ should be set larger than $\alpha_2$ to penalize motions that approach obstacles more heavily.  In Algorithm~\ref{ourMethod} (line~5), the CDF value $\min_{\mathbf{p}'} f_c(\mathcal{P}, \mathbf{q}_t)$ is obtained by evaluating the distance between the current configuration $\mathbf{q}_t$ and all obstacle points in the point cloud $\mathcal{P}$, selecting the minimal distance, and computing the corresponding gradient $\nabla_{\mathbf{q}} f_c(\mathcal{P}, \mathbf{q}_t)$. Then, we sample $N$ control inputs, compute the corresponding next states $\mathbf{q}_{t+1}$, evaluate their costs, and update the MPPI policy accordingly. We apply the mean of the updated distribution as the control input for the next step, and finally project it onto $\mathcal{U}_{\text{step}}(\mathbf{q}_t)$ to ensure that $\mathbf{q}_{t+1}$ remains within the joint limits.

\section{Numerical Simulations}

In this section, we report simulation results on a two-link robot and a Franka Emika robot to evaluate the proposed method. The experiments are designed to (i) demonstrate the necessity of updating both the mean and covariance of the sampling distribution, (ii) show that naïve sampling without distribution updates is insufficient, (iii) assess the effectiveness of our cost-function design, and (iv) compare our approach against representative baselines. Finally, we provide a qualitative analysis to further validate the method. All experiments were performed on a workstation with an AMD Ryzen 7000 CPU and an NVIDIA RTX 4070 GPU.

\begin{algorithm}[t]
    \renewcommand{\algorithmicrequire}{\textbf{Input:}}
    \renewcommand{\algorithmicensure}{\textbf{Output:}}
    \caption{One-step MPPI for Configuration Space Motion Planning}
    \label{ourMethod}
    \begin{algorithmic}[1]
        \REQUIRE Initial configuration $\mathbf{q}_s$; goal configuration $\mathbf{q}_f$; obstacle point cloud $\mathcal{P}$; joint limits $(\mathbf{q}_{\min}, \mathbf{q}_{\max})$; time step $\Delta t$; number of samples $N$; angle-based cost weights $(\alpha_1, \alpha_2)$; obstacle threshold distance $d_{act}$; MPPI parameters.
        \ENSURE Collision-free joint trajectory $\mathbf{q}_{0:T}$

        \STATE Initialize $\mathbf{q}_0 \leftarrow \mathbf{q}_s$
        \STATE Initialize Gaussian policy $\pi_{0} = \mathcal{N}({\boldsymbol{\mu}}_0, {\boldsymbol{\Sigma}}_0)$

        \FOR{$t = 0$ \TO $T-1$}
            \STATE Sample $N$ control inputs $\{\mathbf{u}_i\}_{i=1}^N \sim \pi_{t}$ \COMMENT{one-step ($H = 1$)}
            \STATE Compute CDF value $\min_{\mathbf{p}'} f_c(\mathcal{P}, \mathbf{q}_t)$ and its gradient $\nabla_{\mathbf{q}} f_c(\mathcal{P}, \mathbf{q}_t)$
            \FOR{$i = 1$ \TO $N$}
                \STATE Predict $\mathbf{q}_{t+1}^{(i)} = \mathbf{q}_t + \Delta t \cdot \mathbf{u}_i$
                \STATE Compute angles $\theta_1^{(i)}$, $\theta_2^{(i)}$ using Eqs.~\ref{obs_angle}, \ref{obs_cost_1}, and \ref{obs_cost_2}.
                \STATE Evaluate cost $c^{(i)} = \alpha_1 \theta_1^{(i)} + \alpha_2 \theta_2^{(i)}$
            \ENDFOR
            \STATE Update policy parameters ${\boldsymbol{\mu}}_{t+1}$, ${\boldsymbol{\Sigma}}_{t+1}$ using Eq.~\ref{mean_update} and Eq.~\ref{cov_update}
            \STATE Set the next control input $\mathbf{u}_{\mathrm{des}} \leftarrow {\boldsymbol{\mu}}_{t+1}$
            \STATE Calculate the next state $\mathbf{q}_{t+1}\gets \mathbf{q}_t+\Delta t\,\mathrm{proj}(\mathbf{u}_{\text{des}})$

            \IF{$\lVert \mathbf{q}_{t+1} - \mathbf{q}_f \rVert < \mathrm{allow\_range}$}
                \STATE \textbf{break}
            \ENDIF
        \ENDFOR
    \end{algorithmic}
\end{algorithm}

\subsection{2-DoF Two-Link Robot}
We first evaluate the proposed method on a 2D two-link robot with link lengths $l_1 = l_2 = 2$ and joint limits $q \in [-\pi, \pi]$ for each joint. The hyperparameters used in this experiment are summarized in Table~\ref{tab:hyper2d}. Two circular obstacles are placed at positions $(2.3, -2.3)$ and $(0.0, 2.45)$, each with a radius of $0.3$. The start configuration is $\mathbf{q}_s = [2.1, 1.2]$, and the goal configurations are $\mathbf{q}_f = [-2.1, -0.9]$ and $[-0.5, 0.0]$, and the control time step is $\Delta t = 0.01\,\mathrm{s}$.

\begin{figure*}[t]
    \centering
    \includegraphics[width=\textwidth]{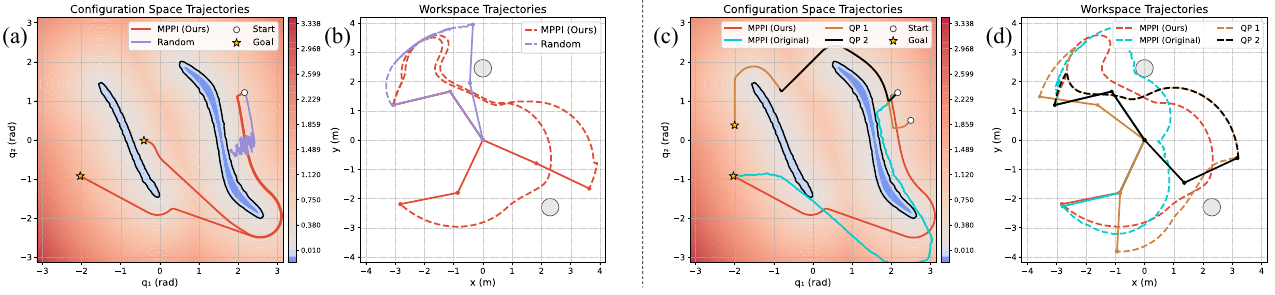}
    \caption{Comparison of different methods in both configuration space and workspace trajectories. The background color encodes the CDF value.
    \textbf{(a)--(b)}: Trajectories from the same start configuration but different goals, used to validate the effectiveness of our cost function design. 
    A purely random baseline, which samples controls independently at each step without policy updates, highlights the necessity of distribution updates in MPPI. 
    \textbf{(c)--(d)}: Comparison with MPPI and an optimization-based method. 
    For QP2 (black), the start and goal are identical to those used in MPPI (Original) and MPPI (Ours), but the trajectory converges to a local optimum and fails to progress. 
    In contrast, QP1, with different start and goal states, successfully reaches the target.}
    \label{Exp_1_trajectories}
\end{figure*}

To highlight the necessity of distribution updates, we consider the following comparison: \textbf{Random Sampling}: At each time step, $N$ random control inputs are sampled, the next configuration is computed, and the state with the lowest cost (Eq.~\ref{cost_our_method}) is selected.
In contrast, our method updates the sampling distribution as described in Algorithm~\ref{ourMethod}.

The results are shown in Fig.~\ref{Exp_1_trajectories}((a)--(b)). 
For random sampling, since the sampling distribution is not updated using information from previous steps, the trajectory may initially move away from obstacles but eventually collides. 
In contrast, by updating the sampling distribution over time, our method progressively biases samples toward safer, lower-cost actions, enabling reliable obstacle avoidance and producing higher-quality trajectories. 
These results support the necessity of distribution updates under the cost in Eq.~\ref{cost_our_method}.

We further provide a comprehensive comparison with other baseline methods: the original MPPI~\cite{sampling_mpc} and the QP-based method from~\cite{cdf}, solved using IPOPT~\cite{ipopt}. The start and goal configurations are the same as in the previous experiment, with an additional test case for the QP method (labeled QP~1), where the start is $[2.5, 0.5]$ and the goal is $[-2.0, 0.3]$. The cost function used in the 2-DoF MPPI experiment is defined as
\begin{equation}
    c_{x, u} = \alpha_g c_{\text{goal}} + \alpha_c c_{\text{col}} + \alpha_j c_{\text{joint}} + \alpha_s c_{\text{stay}}. \label{cost_mpc_2d}
\end{equation}
where $c_{\text{goal}}$ denotes the terminal cost measuring the error between the predicted final configuration and the target configuration $\mathbf{q}_f$, while $c_{\text{stay}}$ penalizes trajectories that remain stationary over time~\cite{sdf2_li2024representing}.

\begin{table}[t]
\centering
\caption{Comparison of Different Methods in the 2D Two-Link Robot}
\label{tab:method_comparison_1}
\begin{tabular}{lccc}
\hline
Method            & Success Rate (\%) & Path Length & Avg. Steps \\
\hline
MPPI (Ours)         & 99.6             & 5.17        & 213 \\
MPPI (Original) \cite{sampling_mpc}   & 71.2              & 4.40        & 155 \\
QP (IPOPT) \cite{cdf}        & 64.1              & 3.75        & 107 \\
\hline
\end{tabular}
\end{table}

The trajectories in configuration space are shown in Fig.~\ref{Exp_1_trajectories}((c)--(d)). The QP-based method~\cite{cdf} occasionally fails to reach the goal for certain start–goal pairs, as the control input oscillates between two states, leading to stagnation and convergence to a local optimum (QP~2, with the same start and goal as MPPI (Ours) and MPPI (Original)). For comparison, in QP~1 with a different start–goal pair, the method successfully reaches the goal. The original MPPI produces trajectories that often run close to obstacles and joint limits, and the resulting paths are generally less smooth.

We also compare the methods using three evaluation metrics: success rate, average number of steps to reach the goal, and average path length, where the path length is defined as
\begin{equation}
L(\mathbf{q}_{0:T}) = \sum_{t=0}^{T-1} \bigl\| \mathbf{q}_{t+1} - \mathbf{q}_{t} \bigr\|_{2},
\label{eq:path-len}
\end{equation}
measured in radians in the joint space.
All methods are tested under the same random seed with identical sampling rules for start and goal states, repeated 500 times. Each trial is terminated when the goal is reached or when a predefined maximum number of steps is exceeded. The results are summarized in Table.~\ref{tab:method_comparison_1}. Our proposed method achieves the highest success rate. Although it requires more steps and produces slightly longer paths, this can be attributed to its ability to avoid local optima and handle longer trajectories effectively. The failure cases mainly occur when the start configuration is very close to an obstacle, causing collisions before the distribution has been sufficiently updated. In contrast, the original MPPI may still result in collisions with obstacles or violations of joint limits due to the difficulty of appropriately balancing the weights among different cost terms. 
\begin{figure}[t]
    \centering
    \includegraphics[width=\linewidth]{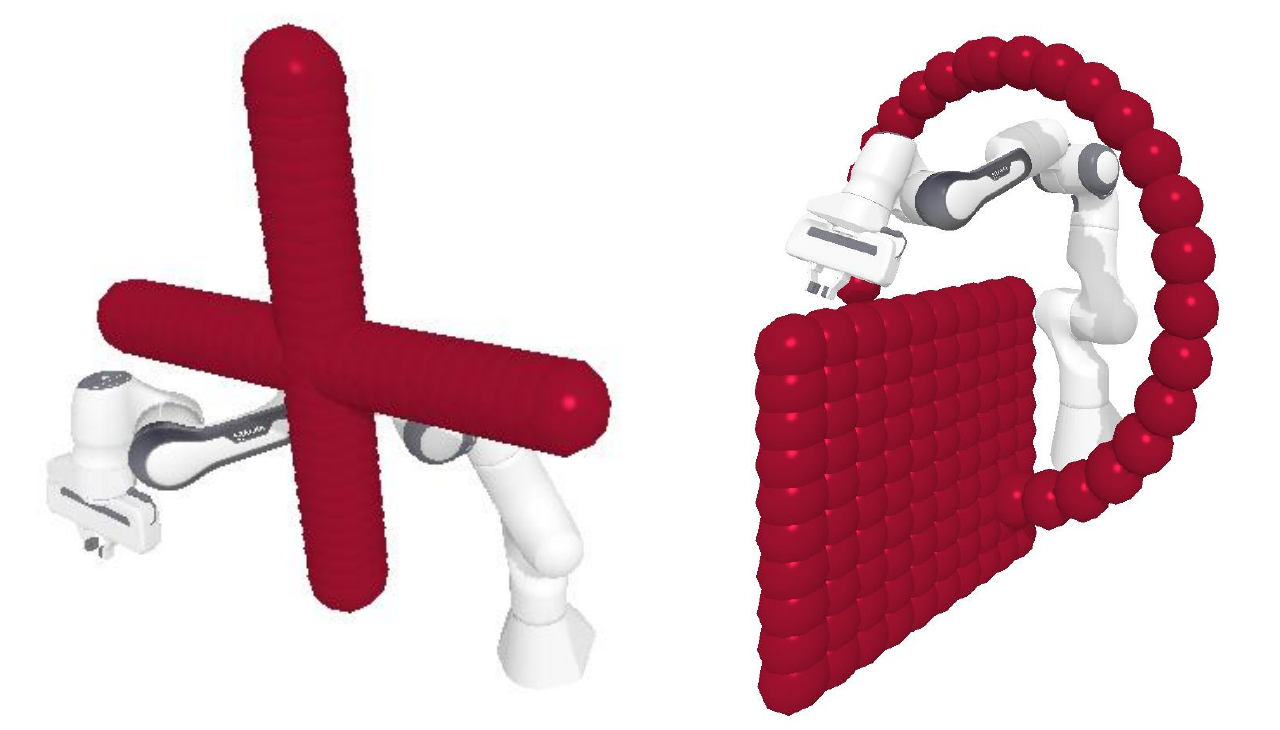}
    \caption{Experimental setups for motion planning with the 7-DoF Franka robot. 
    \textbf{Left}: Scenario 1 with 60 cross-shaped obstacles. The start and goal configurations are fixed, and the obstacle is shifted in each trial such that the linear input between start and goal results in collisions. 
    \textbf{Right}: Scenario 2 with 123 obstacles composed of wall and ring structures. The start and goal configurations are randomly sampled, and their linear input also results in collisions.}
    \label{exp2_scene}
\end{figure}

\begin{figure*}[t]
    \centering
    \includegraphics[width=\textwidth]{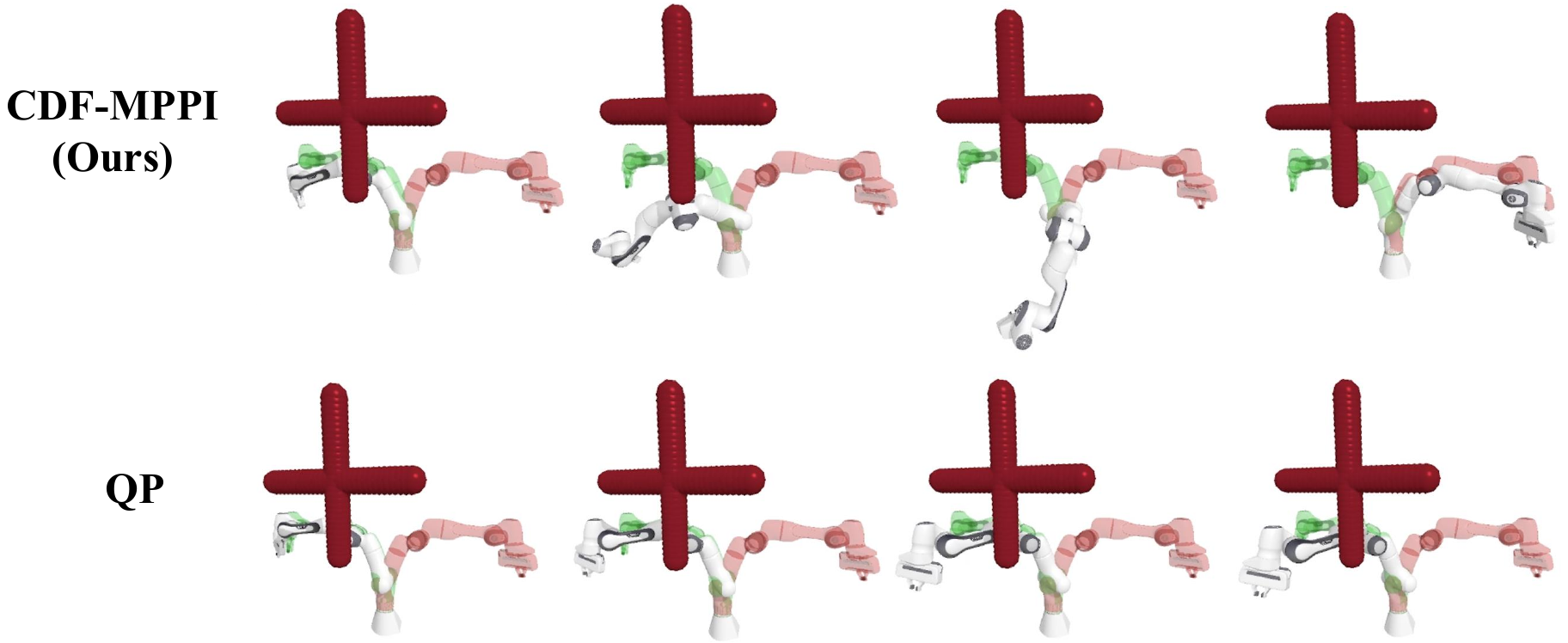}
    \caption{Comparison between the QP method and MPPI (ours) in Scenario~1. The green marker indicates the start posture, while the red marker indicates the goal posture.
    \textbf{Top}: Our method successfully avoids the cross obstacle and reaches the goal. 
    \textbf{Bottom}: The QP method becomes trapped near the obstacle, exhibiting oscillatory behavior around it and failing to make further progress due to local optimality.}
    \label{exp2_franka}
\end{figure*}

\subsection{7-DoF Franka Robot}
In this section, we compare our method with the QP approach from \cite{cdf} under two scenarios (Table~\ref{tab:hyper7d}).  

In the first scenario, shown in Fig.~\ref{exp2_scene}(Left), we follow the experimental setup of \cite{sdf2_li2024representing}. The start and goal configurations are fixed, while the cross-shaped obstacle is shifted in each trial. The robot is commanded using incremental steps of the form
\begin{equation}
    \mathbf{u} = k(\mathbf{q}_f - \mathbf{q}_s). \label{linear_dynamic}
\end{equation}
where $k$ is a small positive coefficient. We refer to this update rule as the \emph{linear dynamics} model. This setup ensures that, across 100 trials, the linear input between the start and goal consistently leads to collisions with the obstacle.  In the second scenario, shown in Fig.~\ref{exp2_scene}(Right), both the start and goal configurations are randomly sampled. Again, 100 trials are conducted, and for each start–goal pair the linear input defined in Eq.~\ref{linear_dynamic} also results in collisions.

The results are summarized in Table~\ref{tab:method_comparison}. Our method achieves consistently high success rates. The path length results indicate that, for the QP method, shorter trajectories are more likely to become trapped in local optima. An example of the QP method being trapped in a local optimum, leading to oscillatory inputs, is illustrated in Fig.~\ref{exp2_franka}. In terms of computational efficiency, our approach achieves an average of 776~Hz in Scenario~1 (with 60 obstacles) and 718~Hz in Scenario~2 (with 123 obstacles). The remaining failures in Scenario~2 are mainly caused by CDF estimation errors when the robot is in near-obstacle configurations, where it is extremely close to obstacles. In such cases, inaccurate distance estimates can lead to unreliable cost evaluation and consequently suboptimal action selection. This limitation stems from the current CDF accuracy rather than the sampling framework itself, and improving CDF fidelity and robustness near obstacles is an important direction for future work. In addition, some failure cases occur when the start configuration is extremely close to an obstacle, leaving insufficient time for the MPPI sampling distribution to adapt before the first few actions are executed, which can result in early collisions, similar to the 2D case.

We further compare the computation frequency with the MPPI method from cuRobo \cite{mppi_hz}. We do not directly compare our method with the original MPPI in the configuration space, primarily because it is challenging to design a suitable cost function. The original MPPI framework is tailored for end-effector tracking, where multiple joint configurations correspond to the same end-effector pose due to the redundancy of the 7-DoF Franka robot. However, when the terminal cost is defined in the configuration space, the solution becomes unique, which makes it difficult for MPPI to converge. Therefore, we only compare the control frequency in this case. Specifically, we provide MPPI with the end-effector poses corresponding to the target joint configurations as its input. Under this setup, MPPI achieves an average of \textbf{61 Hz} in Scenario~1 and \textbf{28 Hz} in Scenario~2. As the number of obstacles increases, collision-checking time also grows. In contrast, \cite{mppi_hz} reported that MPPI can reach up to 500 Hz on an RTX 4090 when the number of obstacles is relatively small. Compared to this, our method sustains significantly higher control frequencies while maintaining near-perfect success rates. Consequently, the proposed method offers a practical advantage for real-time deployment in cluttered scenes, where both high update rates and high success rates are required.



\begin{table}[t]
\centering
\caption{Comparison of Different Methods in the 7-DoF Franka Robot (Two Scenarios)}
\label{tab:method_comparison}
\setlength{\tabcolsep}{3.5pt} 
\renewcommand{\arraystretch}{0.9} 
\begin{tabular}{lccc|ccc}
\hline
 & \multicolumn{3}{c|}{Scenario 1} & \multicolumn{3}{c}{Scenario 2} \\
Method & Succ. (\%) & Len. & Hz & Succ. (\%) & Len. & Hz \\
\hline
MPPI (Ours)       & 100   & 4.57 & 776(59) & 86  & 7.85 & 718(54) \\
QP (IPOPT) \cite{cdf} & 13  & 4.30 & 222(30) & 14 & 6.74 & 228(24) \\
\hline
\end{tabular}
\end{table}

\section{Conclusion}
We have proposed a method that integrates the CDF representation with MPPI. Our approach effectively leverages the complementary strengths of both the CDF representation and MPPI.

The proposed method achieves control frequencies exceeding 750~Hz with consistently high success rates, significantly outperforming the QP baseline, which is prone to local optima. Compared to the standard MPPI framework, our approach simplifies the cost function, unifies its units, and exploits the gradient information provided by the CDF representation to shorten the prediction horizon, thereby attaining substantially faster control frequencies.  

Future work will focus on further optimizing the CDF representation. One direction is to extend the CDF representation beyond point-based sampling in the workspace. In our current experiments, we approximate obstacles using the centers of bounding spheres; a promising extension would allow the CDF representation to compute the shortest distance in configuration space to entire spherical regions. Another direction is to improve robustness near obstacles. When the CDF values are small—i.e., when the manipulator is close to obstacles—collisions can occur due to the highly non-convex nature of the configuration space. Developing strategies that enable safe initialization and reliable avoidance in such regions is an important avenue for future research.

\section{Appendix}
The QP formulation used in this paper follows \cite{cdf}:
\begin{equation}
\mathbf{u}_k^{*} = \arg\min_{\mathbf{q},u} \; e(\mathbf{q}_k)^\top H e(\mathbf{q}_k) + \mathbf{u}_k^\top R \mathbf{u}_k,
\end{equation}
subject to
\begin{align}
\mathbf{q}_{k+1} &= A \mathbf{q}_k + B \mathbf{u}_k, \\
\mathbf{q}_k &\in \mathcal{Q}, \; \mathbf{u}_k \in \mathcal{U}, \\
- \nabla_q f_c(\mathbf{p},\mathbf{q})\, \mathbf{u}_k \Delta t &\leq \ln\!\big(f_c(\mathbf{p},\mathbf{q}) + \gamma \big).
\end{align}
where $\mathbf{q}_k$ and $\mathbf{u}_k$ denote the state and control input at time step $k$, 
$H$ and $R$ are positive definite weighting matrices for tracking error and control effort, 
$e(\mathbf{q}_k) = \mathbf{q}_k - \mathbf{q}_f$ is the configuration error, 
$f_c(\mathbf{p},\mathbf{q})$ is the CDF, 
$\Delta t$ is the time step, $\gamma$ is a safety margin, and $\mathcal{Q}, \mathcal{U}$ are admissible sets of state and control.

For MPPI, the cost function is provided in more detail in Eq.~\ref{cost_mpc_2d}:
\begin{align}
c_{\text{goal}} &= \lVert \mathbf{q}_H - \mathbf{q}_f \rVert_2, \\
c_{\text{col}} &= \sum_{h=1}^H \max(0, -d_h), \\
c_{\text{joint}} &= \sum_{h=1}^H \lVert \max(0, \mathbf{q}_{\min} - \mathbf{q}_h) + \max(0, \mathbf{q}_h - \mathbf{q}_{\max}) \rVert_2^2, \\
c_{\text{stay}} &= \frac{1}{\lVert \mathbf{q}_{H} - \mathbf{q}_{0} \rVert_2 + \varepsilon}.
\end{align}
Here, $\mathbf{q}_h$ denotes the configuration of a sampled trajectory at prediction step $h$ within the horizon $H$.
The term $d_h$ represents the signed distance evaluated by the CDF function at step $h$, where negative values indicate collision with external obstacles. 
\subsection{Hyperparameters}
The hyperparameters used in our experiments for the 2D two-link robot are summarized in Table~\ref{tab:hyper2d}, 
and those for the 7-DoF Franka robot are listed in Table~\ref{tab:hyper7d}. 
The MPPI hyperparameters follow \cite{mppi_hz}, while the QP settings are taken from the official implementation provided in the GitHub repository of \cite{cdf}. 
For the scenario illustrated in Fig.~\ref{exp2_scene} (Left), the start and goal configurations are 
$q_0 = [-1.57,\, 0.40,\, 0.00,\, -1.2708,\, 0.00,\, 1.8675,\, 0.00]^\top$ 
and 
$q_f = [\,1.57,\, 0.40,\, 0.00,\, -1.2708,\, 0.00,\, 1.8675,\, 0.00]^\top$, respectively.
The code and models are publicly available at \url{https://github.com/Rin-Li/cdf_mppi}.

\begin{table}
\centering
\caption{Hyperparameters used in the experiments for 2D two-link robot}
\label{tab:hyper2d}
\setlength{\tabcolsep}{6pt}
\renewcommand{\arraystretch}{1.0}
\begin{tabular}{lll}
\hline
\textbf{Method} & \textbf{Parameter} & \textbf{Value} \\
\hline
QP 
& $H$ & $\mathrm{diag}(100, 35)$ \\
& $R$ & $\mathrm{diag}(0.01, 0.01)$ \\
& $\mathcal{Q}$ & $[-\pi, \pi]$ \\
& $\mathcal{U}$ & $[-3, 3]$ \\
& $\gamma$ & $0.6$ \\
\hline
MPPI 
& $\alpha_g$ & $10$ \\
& $\alpha_c$ & $100$ \\
& $\alpha_j$ & $100$ \\
& $\alpha_s$ & $10$ \\
& Number of samples $N$ & $200$ \\
& Horizon $H$ & $50$ \\
& $\beta$ & $2.0$ \\
& $\gamma$ & $1.0$ \\
& $\alpha_{\boldsymbol{\mu}}$ & $0.5$ \\
& $\alpha_{\boldsymbol{\Sigma}}$ & $0.3$ \\
\hline
Ours (Algorithm~\ref{ourMethod}) 
& $\alpha_1$ & $20$ \\
& $\alpha_2$ & $10$ \\
& $d_{\text{act}}$ & $0.5$ \\
& $N$ & $200$ \\
& $\beta$ & $1.0$ \\
& $\alpha_{\boldsymbol{\mu}}$ & $0.5$ \\
& $\alpha_{\boldsymbol{\Sigma}}$ & $0.5$ \\
\hline
\end{tabular}
\end{table}

\begin{table}
\centering
\caption{Hyperparameters used in the experiments for 7-DoF Franka robot}
\label{tab:hyper7d}
\setlength{\tabcolsep}{6pt}
\renewcommand{\arraystretch}{1.0}
\begin{tabular}{llp{4cm}}  
\hline
\textbf{Method} & \textbf{Parameter} & \textbf{Value} \\
\hline
QP 
& $H$ & $\mathrm{diag}(150,\,190,\,80,\,70,\,70,\,90,\,100)$ \\
& $R$ & $0.01 \cdot I_7$ \\
& $\mathcal{Q}$ & Standard joint limits of the Franka Panda robot \\
& $\mathcal{U}$ & $[-2.7,\, 2.7]$ \\
& $\gamma$ & $0.6$ \\
\hline
Ours (Algorithm~\ref{ourMethod}) 
& $\alpha_1$ & $20$ \\
& $\alpha_2$ & $10$ \\
& $d_{\text{act}}$ & $1.0$ \\
& $N$ & $200$ \\
& $\beta$ & $1.0$ \\
& $\alpha_{\boldsymbol{\mu}}$ & $0.5$ \\
& $\alpha_{\boldsymbol{\Sigma}}$ & $0.5$ \\
\hline
\end{tabular}
\end{table}

\bibliographystyle{IEEEtran}
\bibliography{references}

@inproceedings{sampling_mpc,
  title={Storm: An integrated framework for fast joint-space model-predictive control for reactive manipulation},
  author={Bhardwaj, Mohak and Sundaralingam, Balakumar and Mousavian, Arsalan and Ratliff, Nathan D and Fox, Dieter and Ramos, Fabio and Boots, Byron},
  booktitle={Conference on Robot Learning},
  pages={750--759},
  year={2022},
  organization={PMLR}
}

@INPROCEEDINGS{cdf,
                      author = {Li, Yiming and Chi, Xuemin and Razmjoo, Amirreza and Calinon, Sylvain},
                    projects = {Idiap},
                       title = {Configuration Space Distance Fields for Manipulation Planning},
                   booktitle = {Robotics: Science and Systems (RSS), 2024},
                        year = {2024}
}

@article{ipopt,
  title={On the implementation of an interior-point filter line-search algorithm for large-scale nonlinear programming},
  author={W{\"a}chter, Andreas and Biegler, Lorenz T},
  journal={Mathematical programming},
  volume={106},
  number={1},
  pages={25--57},
  year={2006},
  publisher={Springer}
}

@article{intro_1,
  title={Motion planning with sequential convex optimization and convex collision checking},
  author={Schulman, John and Duan, Yan and Ho, Jonathan and Lee, Alex and Awwal, Ibrahim and Bradlow, Henry and Pan, Jia and Patil, Sachin and Goldberg, Ken and Abbeel, Pieter},
  journal={The International Journal of Robotics Research},
  volume={33},
  number={9},
  pages={1251--1270},
  year={2014},
  publisher={Sage Publications Sage UK: London, England}
}

@inproceedings{intro_2,
  title={CHOMP: Gradient optimization techniques for efficient motion planning},
  author={Ratliff, Nathan and Zucker, Matt and Bagnell, J Andrew and Srinivasa, Siddhartha},
  booktitle={2009 IEEE international conference on robotics and automation},
  pages={489--494},
  year={2009},
  organization={IEEE}
}

@article{sdf1_koptev2022neural,
  title={Neural joint space implicit signed distance functions for reactive robot manipulator control},
  author={Koptev, Mikhail and Figueroa, Nadia and Billard, Aude},
  journal={IEEE Robotics and Automation Letters},
  volume={8},
  number={2},
  pages={480--487},
  year={2022},
  publisher={IEEE}
}

@inproceedings{sdf2_li2024representing,
  title={Representing robot geometry as distance fields: Applications to whole-body manipulation},
  author={Li, Yiming and Zhang, Yan and Razmjoo, Amirreza and Calinon, Sylvain},
  booktitle={2024 IEEE International Conference on Robotics and Automation (ICRA)},
  pages={15351--15357},
  year={2024},
  organization={IEEE}
}

@article{gradient_prob,
  title={Reactive collision-free motion generation in joint space via dynamical systems and sampling-based MPC},
  author={Koptev M, Figueroa N, Billard A.},
  journal={The International Journal of Robotics Research},
  volume={43},
  number={13},
  pages={2049--2069},
  year={2024},
  doi={10.1177/02783649241246557}
}

@inproceedings{mppi,
  title={Information theoretic MPC for model-based reinforcement learning},
  author={Williams, Grady and Wagener, Nolan and Goldfain, Brian and Drews, Paul and Rehg, James M and Boots, Byron and Theodorou, Evangelos A},
  booktitle={2017 IEEE international conference on robotics and automation (ICRA)},
  pages={1714--1721},
  year={2017},
  organization={IEEE}
}

@ARTICLE{GJK,
  author={Gilbert, E.G. and Johnson, D.W. and Keerthi, S.S.},
  journal={IEEE Journal on Robotics and Automation}, 
  title={A fast procedure for computing the distance between complex objects in three-dimensional space}, 
  year={1988},
  volume={4},
  number={2},
  pages={193-203},
  doi={10.1109/56.2083}}

@article{RRT,
  title={Rapidly-exploring random trees: A new tool for path planning},
  author={LaValle, Steven},
  journal={Research Report 9811},
  year={1998},
  publisher={Department of Computer Science, Iowa State University}
}

@ARTICLE{PRM,
  author={Kavraki, L.E. and Svestka, P. and Latombe, J.-C. and Overmars, M.H.},
  journal={IEEE Transactions on Robotics and Automation}, 
  title={Probabilistic roadmaps for path planning in high-dimensional configuration spaces}, 
  year={1996},
  volume={12},
  number={4},
  pages={566-580},
  doi={10.1109/70.508439}}

@INPROCEEDINGS{MPC_1,
  author={Tassa, Yuval and Mansard, Nicolas and Todorov, Emo},
  booktitle={2014 IEEE International Conference on Robotics and Automation (ICRA)}, 
  title={Control-limited differential dynamic programming}, 
  year={2014},
  volume={},
  number={},
  pages={1168-1175},
  doi={10.1109/ICRA.2014.6907001}}

@article{MPC_2,
  title={Model predictive path integral control: From theory to parallel computation},
  author={Williams, Grady and Aldrich, Andrew and Theodorou, Evangelos A},
  journal={Journal of Guidance, Control, and Dynamics},
  volume={40},
  number={2},
  pages={344--357},
  year={2017},
  publisher={American Institute of Aeronautics and Astronautics}
}

@article{mppi_hz,
  title={curobo: Parallelized collision-free minimum-jerk robot motion generation},
  author={Sundaralingam, Balakumar and Hari, Siva Kumar Sastry and Fishman, Adam and Garrett, Caelan and Van Wyk, Karl and Blukis, Valts and Millane, Alexander and Oleynikova, Helen and Handa, Ankur and Ramos, Fabio and others},
  journal={arXiv preprint arXiv:2310.17274},
  year={2023}
}

\end{document}